%% file: root.tex
\documentclass[letterpaper, 10 pt, conference]{ieeeconf}  % Comment this line out if you need a4paper

\IEEEoverridecommandlockouts                              % This command is only needed if 
\usepackage{graphics} % for pdf, bitmapped graphics files
\usepackage{epsfig} % for postscript graphics files
\usepackage{times} % assumes new font selection scheme installed
\usepackage{amsmath} % assumes amsmath package installed
\usepackage{amssymb}  % assumes amsmath package installed
\usepackage{cite}
\usepackage{balance}
\usepackage{amsfonts}
\usepackage{algorithmic}
\usepackage{booktabs}
\usepackage{graphicx}
\usepackage{textcomp}
\usepackage{xcolor}
\usepackage{etoolbox}
\usepackage{subcaption}
\usepackage{multirow}
\usepackage{multicol}
\usepackage{tabularx}
\usepackage{bbding}
\usepackage{array}
\usepackage{makecell}
\usepackage{etex}
\usepackage{hyperref}
\title{\LARGE \bf
GSDrive: Reinforcing Driving Policies by Multi-mode Future Trajectory Probing with 3D Gaussian Splatting Environment}
\author{Ziang Guo$^{\dagger}$, Chen Min$^{\circ*}$, Xuefeng Zhang$^{\ddag}$, Yixiao Zhou$^{\S}$, \\ Shuo Wang$^{\circ}$, Sifa Zheng$^{\ddag}$, Dzmitry Tsetserukou$^{\dagger}$, and Zufeng Zhang$^{\ddag*}$
\thanks{$^{*}$ are the corresponding authors.}
\thanks{$^{\dagger}$Ziang Guo and Dzmitry Tsetserukou are with the Intelligent Space Robotics Laboratory, Skolkovo Institute of Science and Technology, Moscow, Russia., {\tt\small \{ziang.guo, d.tsetserukou\}@skoltech.ru}}%
\thanks{$^{\circ}$Chen Min and Shuo Wang are with the Research Center for Intelligent Computing Systems, SKLP, Institute of Computing Technology, Chinese Academy of Sciences, Beijing, China., {\tt\small \{mincheng, wangshuo24z\}@ict.ac.cn}}%
\thanks{$^{\S}$Yixiao Zhou is with the Department of Electrical and Electronic Engineering, The University of Hong Kong, China., {\tt\small u3649491@connect.hku.hk}}%
\thanks{$^{\ddag}$Xuefeng Zhang is with SuZhou Automotive Research Institute, Tsinghua University, China. Zufeng Zhang and Sifa Zheng are with SuZhou Automotive Research Institute, Tsinghua University and the School of Vehicle and Mobility, Tsinghua University, China., {\tt\small osu\_zxf@126.com, zhangzufeng@tsari.tsinghua.edu.cn, zsf@tsinghua.edu.cn}}%
}

\begin{document}

\maketitle
\thispagestyle{empty}
\pagestyle{empty}

\begin{abstract}

End-to-end (E2E) autonomous driving aims to directly map sensory observations to driving actions, but its real-world deployment is hindered by evolving data distributions and the high cost of continual annotation. While combining imitation learning (IL) and reinforcement learning (RL) is a common strategy for policy improvement, conventional RL training relies on delayed, event-based rewards, where policies learn only from catastrophic outcomes such as collisions, leading to premature convergence to suboptimal behaviors. To address these limitations, we propose GSDrive, a framework that uses a differentiable 3D Gaussian Splatting (3DGS) environment for future-aware trajectory probing and reward shaping in E2E driving. GSDrive first learns a multi-mode trajectory probe via IL and then uses RL to evaluate multiple candidate futures in the 3DGS environment, converting their simulated returns into dense shaping rewards for policy optimization. This yields a cyclic hybrid IL–RL training loop, where IL supplies structured future priors and RL provides interactive feedback for iterative refinement. Evaluated on the reconstructed nuScenes dataset, our method outperforms other simulation-based RL approaches in closed-loop experiments. Code will be made publicly available at \href{https://github.com/ZionGo6/GSDrive}{\textit{https://github.com/ZionGo6/GSDrive}}.

\end{abstract}

\section{INTRODUCTION}

\input{Intro}

\section{RELATED WORK}

\input{Related}

\section{METHODOLOGY}

\input{Method}

\section{EXPERIMENTS}

\input{Exp}

\section{CONCLUSIONS}

\input{Con}

\bibliographystyle{IEEEtran}
\bibliography{IEEEfull, refs}

\end{document}

%% file: Intro.tex
End-to-end (E2E) driving policies provide an efficient paradigm for mapping real-world observations directly to driving actions \cite{chen2024end}. However, deploying such policies in continuously evolving traffic environments remains challenging \cite{nebot2026era}. As real-world data distributions shift over time, it is impractical to indiscriminately incorporate all newly collected data into E2E policy training due to both the high cost of annotation and the difficulty of maintaining consistent data quality at scale \cite{ye2025iterative}. A common strategy to improve E2E policies is to combine imitation learning (IL) with reinforcement learning (RL). While IL provides strong behavioral priors from expert demonstrations, it is inherently limited by the quality and coverage of the demonstration data. RL, in contrast, enables policy improvement through interaction, but its effectiveness is often hindered by sparse and short-sighted reward signals. In complex driving scenarios, critical failures such as collisions are usually caused by a sequence of suboptimal decisions, making it difficult for conventional RL methods to assign meaningful credit to earlier actions.

To address this limitation, we propose GSDrive, a novel framework that introduces future-aware reward shaping through multi-mode trajectory probing in a 3D Gaussian Splatting (3DGS) environment. The key idea is to evaluate multiple candidate future trajectories and use their simulated outcomes to guide current decision-making. Rather than relying solely on sparse terminal signals, GSDrive provides dense and informative feedback by explicitly estimating the long-term consequences of current actions.

Concretely, we first train a multi-mode trajectory predictor with IL using a flow-matching objective, enabling it to model diverse plausible driving behaviors. During RL, the predicted trajectories are probed in the 3DGS environment to obtain future returns, which are then used for reward shaping. In this way, imitation-derived trajectory priors and RL-based exploration are coupled within a unified learning framework, allowing for policy improvement beyond the demonstrated distribution while retaining structured behavioral guidance.

Our hybrid learning approach allows continuous policy improvement through simulation-based RL, where new unlabeled real-world observations can be incorporated into the 3DGS environment reconstruction without expensive annotation, thus supporting lifelong learning capabilities essential for production autonomous driving systems \cite{zeng2025gaussianupdate}. Our contribution is summarized as follows: 
\begin{itemize}
    \item We introduce a method for constructing future-aware rewards through multi-mode trajectory probing in a 3DGS environment, forming a cyclic hybrid learning process in which IL provides structured trajectory priors and RL supplies dense feedback for iterative policy refinement.
    \item We present an efficient E2E policy refinement strategy by integrating unlabeled real-world data into the 3DGS reconstruction. This allows for continuous, simulation-based policy learning without the bottleneck of exhaustive human annotation.
    \item We propose a flexible learning framework that allows for the calibration of the trade-off between the IL-derived priors and the RL-learned residuals to align with different performance goals.
\end{itemize}

%% file: Related.tex
\subsection{Generative E2E Autonomous Driving}

Recent research has shaped E2E autonomous driving with profound advances \cite{sun2025sparsedrive, li2025end, sima2025centaur, naumann2025data, yang2025uncad, kim2025synad}. Along with the development of generative decision-making, DiffE2E couples hierarchical bidirectional cross attention for multi-sensor feature alignment with a Transformer-based hybrid diffusion-supervision decoder, which enables the model to generate multi-mode future trajectories \cite{zhao2025diffe2e}. DistillDrive proposes a knowledge distillation framework for the isomorphic source planning model, which uses the example output of the multi-mode planning teacher model to supervise the E2E student model. It further combines multi-mode instance imitation, reinforcement learning, and generative modeling to enhance planning-oriented feature learning \cite{yu2025distilldrive}. ARTEMIS integrates autoregressive E2E trajectory planning and Mixture-of-Experts (MoE) mechanisms into E2E planning and captures time dependence through sequence trajectory point generation \cite{feng2025artemis}. However, these generative decision-making paradigms still face challenges in generalization and robustness, as the modeling of complex multi-mode distributions heavily relies on the quality and diversity of offline demonstration data.

\subsection{RL in Autonomous Driving}

RL is a common solution in E2E autonomous driving to improve exploration, safety, and closed-loop robustness. ZTRS presents a zero-imitation E2E autonomous driving framework that replaces expert imitation with trajectory scoring and offline RL. It proposes an exhaustive strategy optimization (EPO) for offline RL, which shows stronger robustness in safety-critical scenarios \cite{li2025ztrs}. More closely related to the generalization of E2E driving policies, ReCogDrive combines VLM reasoning with diffusion planners and introduces DiffGRPO to strengthen the planner, making it go beyond pure IL \cite{li2025recogdrive}. DriveDPO proposes that IL may generate human-like but unsafe trajectories and uses Safety DPO to directly optimize strategy distribution through trajectory-level preference alignment \cite{shang2025drivedpo}. TakeAD uses experts to take over data and then post-optimize end-to-end strategies in the separation scenario with DPO \cite{liu2025takead}. These methods primarily focus on strategy fine-tuning, preference alignment, or value-based selection, while the critical reward feedback used for policy improvement remains under-exploited.

\subsection{Simulation-based Autonomous Driving}

To enhance the supervisory signals, simulation-based E2E autonomous driving methods have been widely discussed. RAD builds a closed-loop RL framework based on 3DGS, which reconstructs a realistic digital driving environment. It allows policies to interact with the environment through rolling drills and to design safety-related rewards to close the open-loop gap between causal confusion and IL \cite{gao2025rad}. ReconDreamer-RL further improves reconstruction-based RL by integrating a video diffusion prior into scene reconstruction \cite{ni2025recondreamer}. Drive\&Gen connects the video generation model with the E2E planner to evaluate the quality of synthetic data through controllable generated videos, analyze the domain gap of the planner, and improve the out-of-distribution generalization ability \cite{wang2025drivegen}. FutureX enhances E2E planning through a latent world model, where future scene representations are rolled out to refine motion plans \cite{lin2025futurex}. Despite these advancements, these methods often rely on sparse, event-triggered reward signals that fail to accurately attribute long-term consequences to specific actions, often leading to suboptimal policy convergence. Based on these insights, we propose designing a more effective E2E strategy optimization framework by closely coupling future trajectories with the physical simulation through multi-mode trajectory probes in the 3DGS environment to provide future-aware rewards.

%% file: Method.tex
\subsection{3DGS Reconstruction}

We leverage VGGT, a multi-view Transformer, to extract geometric features from six surrounding camera images by constraining cross-view correspondence searches to epipolar lines derived from the Essential Matrix, encoding relative camera poses \cite{wang2025vggt}. The decoder predicts Gaussian primitives in a single forward pass, recovering 3D positions via back-projection from predicted depth, along with scale factors, quaternion-based orientation, opacity, and Spherical Harmonics color coefficients, which construct a valid covariance matrix for rendering. Color and depth are computed through front-to-back alpha-blending of Gaussians, supervised by a combined loss that balances photometric accuracy, structural similarity, and LiDAR depth consistency from nuScenes. For on-the-fly rasterization, 3D Gaussians are projected to 2D by transforming their covariance matrices, sorting them by depth, assigning them to image tiles, and blending their contributions in sorted order to produce the final rendered images $\mathbf{\hat{I}}$.

\begin{figure*}[h!]
    \centering
    \includegraphics[width=0.96\linewidth]{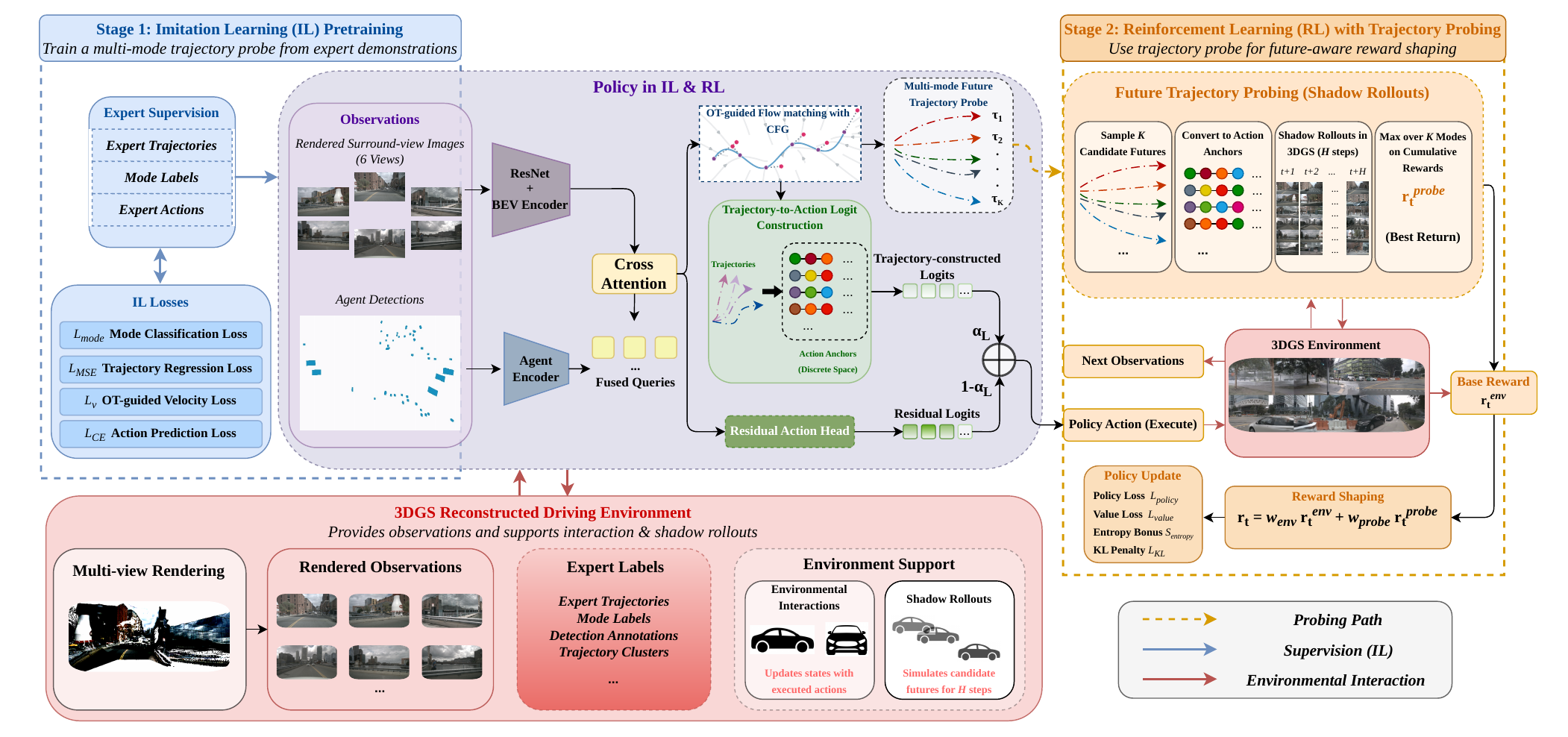}
    \caption{GSDrive pipeline.}
    \label{fig:main}
\vspace{-3mm}
\end{figure*}
\subsection{Flow Matching Trajectory Prediction}

The overall pipeline is shown in Fig. \ref{fig:main}, where the policy obtains 6 rendered images and agent detections as observations from the 3DGS environment. The agent detections are the corresponding detection labels in nuScenes. Based on \cite{philion2020lift}, the images $\mathbf{\hat{I}} \in \mathbb{R}^{6C \times H \times W}$ are compressed into a Bird's-eye view (BEV) space to construct spatial features, while the image semantic features are extracted via ResNet \cite{He_2016_resnet}. Through cross attention blocks, fused queries are directed to a flow matching-based head to obtain trajectory predictions $\mathbf{T}^{N_{\text{modes}} \times N_{\text{points}} \times 2}$ and their mode classifications $\mathbf{C}^{N_{\text{modes}}}$ under Classifier-Free Guidance (CFG) \cite{ho2022classifier}, where $N_{\text{points}} = 6$ indicates the number of trajectory points for the future $3$ seconds, and $N_{\text{modes}}$ represents the number of trajectory modes based on the clustered trajectories in the nuScenes dataset.

In our flow matching head, the goal is to learn a velocity field $\mathcal{V}_\theta$ that transports the distribution of clustered trajectories $p_0 \in \mathcal{P}(\mathcal{T}_0)$ to that of the ground truth trajectories $p_1 \in \mathcal{P}(\mathcal{T}_1)$. We incorporate Optimal Transport (OT) to guide trajectories to minimize the transport cost between probability distributions \cite{kornilov2024optimal}. The optimal coupling $\pi^* \in \Pi(p_0, p_1)$ is defined as:
\begin{equation}
    \pi^* = \arg\min_{\pi \in \Pi(p_0, p_1)} \int_{\mathcal{T}_0 \times \mathcal{T}_1} \|\tau_0 - \tau_1\|^2 \, d\pi(\tau_0, \tau_1),
\end{equation}
where the set of all joint distributions with marginals $p_0$ and $p_1$ is
\begin{equation}
    \Pi(p_0, p_1) = \left\{ \pi \in \mathcal{P}(\mathcal{T}_0 \times \mathcal{T}_1) : \pi(\cdot, \mathcal{T}_1) = p_0, \pi(\mathcal{T}_0, \cdot) = p_1 \right\}.
\end{equation}

Computing the optimal coupling exactly is computationally intensive for large datasets. We therefore employ entropic regularization via the Sinkhorn algorithm \cite{pham2020unbalanced}, which solves the regularized optimal transport problem:
\begin{equation}
    \text{OT}(p_0, p_1) = \min_{\pi \in \Pi(p_0, p_1)} \int \|\tau_0 - \tau_1\|^2 \, d\pi(\tau_0, \tau_1) + \epsilon_1 H^{OT}(\pi),
\end{equation}
where $H^{OT}(\pi)$ is the Shannon entropy of the coupling, and $\epsilon_1 > 0$ is the regularization parameter. The entropic term encourages smoother couplings and enables efficient computation via fixed-point iterations. Upon convergence, the optimal coupling is obtained as:
\begin{equation}
    {P^* = \text{diag}(u^{OT}) \, K^{OT} \, \text{diag}(v^{OT})},
\end{equation}
where $u^{OT}$ and $v^{OT}$ are the scaling vectors. $P^*_{ij}$ represents the optimal transport mass, and $K^{OT}_{ij} = \exp\left(-\frac{\|\tau_0^{(i)} - \tau_1^{(j)}\|^2}{\epsilon_1}\right)$ measures the affinity from the source point $i$ to the target point $j$.

With the computed coupling $P^*_{ij}$, we define the OT-guided path interpolation, which transports clustered trajectories to ground truth trajectories along cost-minimizing paths:
\begin{equation}
    \tau_t^{(i)} = \sum_{j=1}^{N_{\text{traj}}} \frac{P^*_{ij}}{a_i} \left[(1-t^v)\tau_0^{(i)} + t^v \tau_1^{(j)}\right],
\end{equation}
where $N_{\text{traj}}$ is the number of computed trajectories. $a_i = \sum_{j=1}^{N_{\text{traj}}} P^*_{ij}$ is the marginal weight of the source trajectory $i$, $\frac{P^*_{ij}}{a_i}$ is the normalized coupling weight, representing the probability of transitioning from $\tau_0^{(i)}$ to $\tau_1^{(j)}$ under the optimal transport plan. $t^v \in [0, 1]$ is the interpolation time.

The velocity target at time $t$ is defined as the OT-weighted displacement:
\begin{equation}
    \nu_t^{\text{OT}}\left(\tau_t^{(i)} \mid \tau_0^{(i)}, \tau_1^{(j)}\right) = \mathbb{E}_{\tau_1^{(j)} \sim \pi^*(\cdot \mid \tau_0^{(i)})}\left[\tau_1^{(j)} - \tau_0^{(i)}\right].
\end{equation}

Expanding the expectation, we have 
\begin{equation}
    \nu_t^{\text{OT}} = \sum_{j=1}^{N_{\text{traj}}} w_j^{(i)} \left(\tau_1^{(j)} - \tau_0^{(i)}\right),
\end{equation}
where $w_j^{(i)} = \frac{P^*_{ij}}{a_i}$ is the coupling weight, representing the relative importance of target $\tau_1^{(j)}$ in transporting source $\tau_0^{(i)}$.

The OT-weighted velocity loss for training the velocity field $\mathcal{V}_\theta$ is:
\begin{equation}
    L_v = \mathbb{E}_{t^v \sim \mathcal{U}[0,1]} \left[ \sum_{i=1}^{N_{\text{traj}}} w_i \left\| \mathcal{V}_\theta\left(\tau_t^{(i)}, t^v\right) - \nu_t^{\text{OT}} \right\|^2 \right],
\end{equation}
where $w_i$ is the normalized marginal weight. This formulation ensures that $\mathcal{V}_\theta$ learns to approximate the velocity field that transports clustered trajectories to ground truth trajectories along the OT paths, with a heavier emphasis on well-coupled mode pairs during training.

\subsection{Action Logit Construction}

The action logits are constructed based on trajectory predictions $\mathbf{T}$ and mode classifications $\mathbf{C}$. From each point $(x,y)$ in a trajectory, we define a grid space within the spatial boundaries $[x_{min}, x_{max}]$ and $[y_{min}, y_{max}]$. Then the action anchor sets $\mathcal{A}_x$ and $\mathcal{A}_y$ are uniformly spaced sequences within the defined grid space $\mathcal{A}_x = \{a_{x,1}, \dots, a_{x,N_{\text{anchors}}}\}, \quad \mathcal{A}_y = \{a_{y,1}, \dots, a_{y,N_{\text{anchors}}}\}$. The similarity between a trajectory point and an action anchor is computed using an exponential kernel based on the normalized absolute distance in all the trajectory modes $\mathbf{C}$.
\begin{equation}
\begin{split}
    s_{x, (m, n)} = \exp\left( -\frac{|x_m - a_{x,n}|}{\tau_s \cdot \Delta x_{\max}} \right) \\
    s_{y, (m, n)} = \exp\left( -\frac{|y_m - a_{y, n}|}{\tau_s \cdot \Delta y_{\max}} \right),
\end{split}
\end{equation}
where $\tau_s$ is the scaling hyperparameter, $m \in \mathbf{C}$ is the mode index, and $n \in N_{\text{anchors}}$ is the action anchor index.

Then, the per-mode similarity is normalized by Softmax with a temperature of $0.5$, e.g.
\begin{equation}
\log P(n | m)_x = \text{log-softmax}\left( \frac{s_{x, (m, n)}}{0.5} \right).
\end{equation}

The action logits are constructed with the normalized classification weights $w_m$ for mode $m \in \mathbf{C}$, representing the total log-probability across all modes.
\begin{equation}
\begin{split}
    \text{Logit}_x(a_n) = \log \sum_{m} \exp \left( \log P(n | m)_x + \log(w_m) \right) \\
    \text{Logit}_y(a_n) = \log \sum_{m} \exp \left( \log P(n | m)_y + \log(w_m) \right).
\end{split}
\end{equation}

Alongside the constructed action logits from the trajectory, an MLP-based residual action head is used to obtain residual action logits from fused queries directly. The final action logits are combined as
\begin{equation}
\begin{split}
    \textbf{Logits}_x &= \alpha^L \cdot \text{Logit}_x + (1-\alpha^L) \cdot \text{Residual Logit}_x \\
    \textbf{Logits}_y &= \alpha^L \cdot \text{Logit}_y + (1-\alpha^L) \cdot \text{Residual Logit}_y,
\end{split}
\end{equation}
where $\alpha^L$ controls the ratio of the residual logits.

For different training objectives, the ratio of residual action logits and the weights of losses is adjusted. In IL, we assign a large weight to trajectory prediction and a low residual ratio to action generation, targeting the learning of a trajectory probe in the 3DGS environment. In RL, a large weight and a high residual ratio for action generation are set to reinforce the policy through the interaction between the trajectory probe and the 3DGS environment. We introduce more details of each stage in the following sections.

\subsection{IL-learned Trajectory Probe}

In IL, the objectives involve imitating trajectory modes clustered from expert trajectories, as well as directly learning from expert trajectories and expert actions. For trajectory modes, the objective is multi-mode classification.
\begin{equation}
    L_{\text{mode}} = \frac{1}{N_m \cdot C_m} \sum_{i=1}^{N_m} \sum_{j=1}^{C_m} \text{Focal}(p^{m}_{i,j}, y^{m}_{i,j}),
\end{equation}
where the focal loss is 
\begin{equation}
    \text{Focal}(p^{m}_{i,j}, y^{m}_{i,j}) = 
\begin{cases} 
-\zeta (1 - p^{m}_{i,j})^\mu \log(p^{m}_{i,j}) & \text{if } y^{m}_{i,j} = 1 \\
-(1 - \zeta) {p^{m}_{i,j}}^\mu \log(1 - p^{m}_{i,j}) & \text{if } y^{m}_{i,j} = 0.
\end{cases}
\end{equation}
Among the above, $C_m$ and $N_m$ are the numbers of modes and samples within a batch. $p^{m}_{i,j}$ is the sigmoid probability for mode $j$. $y^{m}_{i,j}$ is the $j$-th element of the one-hot encoded target vector for sample $i$. $\zeta$ and $\mu$ are the hyperparameters that manage the imbalance between the modes.

The overall IL objectives are, along with the OT-weighted velocity loss $L_{v}$ for the velocity field prediction, the MSE loss $L_{\text{MSE}}$ for the trajectory regression towards the ground truth trajectory $\mathbf{\hat{T}}$, and the cross entropy loss $L_{\text{CE}}$ for the action logits against the constructed $\mathbf{\hat{Logits}}$ from $\mathbf{\hat{T}}$.
\begin{equation}
\begin{split}
    L_{\textbf{IL}} &= w_{\text{mode}} \cdot L_{\text{mode}} \\ &+ w_{\text{traj}} \cdot \left(L_{\text{MSE}}(\mathbf{T}, \mathbf{\hat{T}}) + L_{v}\right)\\ &+ w_{\text{action}} \cdot L_{\text{CE}}(\mathbf{Logits}, \mathbf{\hat{Logits}}),
\end{split}
\end{equation}
where the weights $w_{\text{mode}}$, $w_{\text{traj}}$, and $w_{\text{action}}$ are for balancing the loss scale. In the IL stage, as a warm start to RL training, the loss is with higher weights of $L_{\text{MSE}}$ and $L_{\text{mode}}$ than $L_{\text{CE}}$ to focus on trajectory prediction.

\subsection{RL with Multi-mode Trajectory Probing}

We formulate the RL training in a Markov Decision Process (MDP) defined as $(\mathcal{S}, \mathcal{A}, \mathcal{T}, \mathcal{R}, \gamma)$. For the state space $\mathcal{S}_t$, it combines camera images $\mathbf{\hat{I}}_t$, agent detection $\mathbf{A}_t$, and camera intrinsic and extrinsic matrices $\mathbf{K}_t$ and $\mathbf{E}_t$ for image projection into $(\mathbf{\hat{I}}_t, \mathbf{A}_t, \mathbf{K}_t, \mathbf{E}_t)$. The action space $\mathcal{A}_t$ is defined by the grid space of trajectory points. The policy outputs logits for a categorical distribution over anchors in the grid space. The action is sampled and then mapped to specific trajectory endpoints. The transition dynamics $\mathcal{T}$ are defined by the transition function $s_{t+1} \sim \mathcal{T}(\cdot | s_t, a_t)$, which is governed by the 3DGS environment physics and the policy action output, where the environment takes the current state and action to simulate the next state $s_{t+1}$ of the world.

The reward function $r_t = \mathcal{R}(s_t, a_t, s_{t+1})$ is formulated as: 
\begin{equation}
    r_t = \overbrace{w_{\text{env}} \cdot r_t^{\text{env}}}^{\text{Base Rewards}} + \overbrace{w_{\text{probe}} \cdot \max_{i \in K}(r_{\text{probe}}(\tau_i))}^{\text{Trajectory Probing}},
\end{equation}
where $w_{\text{env}}, w_{\text{probe}}$ are the weights for scale balancing. $r_t^{\text{env}}$ is the combination of base interactions, including Survival Bonus (reward for not colliding or going off-road), Progress Reward (distance covered along the expert trajectory), Collision Penalty (negative penalty for dynamic and static collisions), and Comfort Penalty (negative penalty based on jerk and acceleration). $r_{\text{probe}}(\tau_i)$ is the trajectory probing reward, where, under the current observation, the flow matching trajectory head is iterated $K$ times to sample $K$ diverse trajectory candidates with varied CFG scales across iterations. A lower guidance scale reduces adherence to conditioning signals and allows the head to explore more diverse mode configurations. Each predicted trajectory is converted into executable action anchors following the action logit construction for environment probing. The probing operates on a shadow copy of the environment states, where the current environment state is stored before any actions are executed. Then, actions are executed for $H$ horizons, computing base rewards at each step. Finally, the environment state is restored after all probing steps.

As an auxiliary reward shaping technique to improve temporal credit assignment in driving scenarios, the probing reward assigned to the current action is computed as the maximum reward obtained among all $K$ probed trajectories, thereby encouraging the policy to learn from the optimal yet viable future.
\begin{equation}
      r_t^{\text{probe}} = \max_{i=1...K} \left[ \sum_{h=0}^{H} \gamma^h r^{\text{env}}(s_{t+h}, a_{t+h}) \right],
\end{equation}
where $H$ represents the probe horizon that the policy looks ahead at during the trajectory probing process. It defines the future time steps of the trajectory that the policy simulates forward in the 3DGS environment to evaluate the quality of the current action. $\gamma^h$ is the discount factor for the probing reward. 

As shown in Fig. \ref{fig:main}, the policy is then shaped by a clipped surrogate objective $\mathcal{L}_{\text{policy}}$, a value loss $\mathcal{L}_{\text{value}}$, and an entropy bonus $\mathcal{S}_{\text{entropy}}$. Having the probability ratio $\eta_t(\theta)$ from the categorical distributions over the action anchors,
\begin{equation}
     \eta_t(\theta) = \frac{\pi_{\theta}(a_t | s_t)}{\pi_{\theta_{\text{old}}}(a_t | s_t)},
\end{equation}
where $\pi_{\theta}$ and $\pi_{\theta_{\text{old}}}$ denote the updated policy and the old policy, the clipped surrogate objective $\mathcal{L}_{\text{policy}}$ is formulated as follows:
\begin{equation}
    \mathcal{L}_{\text{policy}} = -\hat{\mathbb{E}}_t \left[ \min(\eta_t(\theta) \hat{A}_t, \text{clip}(\eta_t(\theta), 1 - \epsilon, 1 + \epsilon) \hat{A}_t) \right],
\end{equation}
where $ \hat{A}_t$ is the estimated advantage derived from rewards. $\epsilon$ is the limitation of the probability ratio $\eta_t(\theta)$.

The value loss is defined as
\begin{equation}
    \mathcal{L}_{\text{value}} = \hat{\mathbb{E}}_t \left[ \left( V_{\theta}(s_t) - Re_t \right)^2 \right],
\end{equation}
where $V_{\theta}$ is the value head. $Re_t$ is the estimated return, calculated as the sum of the Generalized Advantage Estimation (GAE) and the old value estimate.

The entropy bonus $\mathcal{S}_{\text{entropy}}$ is meant to encourage exploration by preventing the policy from becoming too deterministic, which is typically associated with Shannon Entropy $\mathcal{H}$.
\begin{equation}
    \mathcal{S}_{\text{entropy}} = \mathbb{E}_t \left[ \mathcal{H}(\pi_{\theta_{\text{old}}}(\cdot | s_t)) \right].
\end{equation}

Beyond direct reward maximization, monitoring and tuning the level of policy stochasticity is crucial for our multi-mode probing exploration. We therefore augment the RL objective with a KL divergence regularization, which constrains the policy from deviating too sharply from a trajectory probe. The approximation of KL divergence is computed as follows.
\begin{equation}
    \bar{D}_{KL}(\theta_{old} || \theta) \approx \mathbb{E}_t\left[\frac{\pi_\theta(a_t|s_t)}{\pi_{\theta_{old}}(a_t|s_t)} - 1 - \log\frac{\pi_\theta(a_t|s_t)}{\pi_{\theta_{old}}(a_t|s_t)}\right].
\end{equation}

Then, we use an Exponential Moving Average (EMA) to track KL divergence in order to achieve an adaptive KL controller.
\begin{equation}
    \bar{D}_{\text{EMA},t} = \alpha_{\text{KL}} \cdot \bar{D}_{\text{EMA}, t-1} + (1-\alpha_{\text{KL}}) \cdot \bar{D}_t,
\end{equation}
where $\alpha_{\text{KL}}$ is equal to $0.9$, meaning $90\%$ of weight is assigned to history KL divergence and $10\%$ to current KL divergence.

Meanwhile, an adaptive KL penalty mechanism is defined with a penalty coefficient $\kappa_{\text{adaptive}}$ that adapts based on the KL deviation from the target $\bar{D}_{\text{target}}$. Given
\begin{equation}
    \delta_{\text{KL}} = \frac{\bar{D}_{\text{EMA}} - \bar{D}_{\text{target}}}{\bar{D}_{\text{target}}},
\end{equation}
\begin{equation}
    \kappa_{\text{adaptive}} = \begin{cases} 
        \kappa_{\text{base}} \cdot (1 + 2\delta_{\text{KL}}) & \text{if } \bar{D}_{\text{EMA}} > \bar{D}_{\text{target}} \\
        0.8 \cdot \kappa_{\text{base}} & \text{if } \bar{D}_{\text{EMA}} < 0.5 \bar{D}_{\text{target}} \\
        \kappa_{\text{base}} & \text{otherwise},
    \end{cases}
\end{equation}
where $\kappa_{\text{base}}$ and $\bar{D}_{\text{target}}$ are initialized as the learnable parameters.

Furthermore, we construct the KL penalty loss with a proactive formulation of linear and quadratic penalties. Given
\begin{equation}
    \Delta_{\text{KL}} = \max(0, \bar{D}_{\text{EMA}} - 0.5 \bar{D}_{\text{target}}),
\end{equation}
\begin{equation}
 \mathcal{L}_{\text{KL}} = 
\begin{cases} 
\kappa_{\text{adaptive}} (2\Delta_{\text{KL}} + \frac{1}{2}\Delta_{\text{KL}}^2) & \text{if } \bar{D}_{\text{EMA}} > 0.25 \bar{D}_{\text{target}} \\
0 & \text{otherwise},
\end{cases}
\end{equation}
targeting a differentiable KL divergence regularization that complements our flexible learning framework. The final RL objectives are computed as:
\begin{equation}
    L_{\textbf{RL}} = \mathcal{L}_{\text{policy}} + c_1 \cdot \mathcal{L}_{\text{value}} - c_2 \cdot \mathcal{S}_{\text{entropy}}[\pi_\theta(\cdot|s_t)] + c_3 \cdot \mathcal{L}_{\text{KL}},
\end{equation}
where $c_1$, $c_2$, and $c_3$ are the balancing coefficients for value loss, entropy bonus, and KL divergence loss.

In our overall pipeline, we leverage OT-guided flow matching with CFG scales for diverse future trajectory prediction and residual action logits to support both IL for stable initialization and RL for adaptive refinement. To improve temporal credit assignment in RL, we introduce a future-aware trajectory probing mechanism that evaluates candidate trajectories through environment rollouts, providing dense reward shaping without requiring additional learned critics. Through KL divergence regularization, the policy updates are monitored and stabilized during exploration.

%% file: Exp.tex
\begin{table*}[h]
% \vspace{0.13cm}
\centering
\caption{Closed-loop metrics comparison on all the methods.}
\resizebox{0.88\textwidth}{!}{
\begin{tabular}{c|c c c c c c c}
\toprule
     Method & ER $\uparrow$ & DS $\uparrow$ & MA $\downarrow$ & LC $\uparrow$ & MAJ $\downarrow$ & MSA $\downarrow$ & CR $\downarrow$ \\ \hline
    PPO \cite{schulman2017proximal} & $26.76$ & $10.92$ & $2.65$ & $0.60$ & $1.08$ & $0.11$ & $0.39$ \\
    Skill-Critic \cite{hao2024skill} & $30.38$ & $10.68$ & $2.78$ & $1.87$ & $1.12$ & $0.08$ & $0.28$ \\
    RLIR \cite{ye2025reinforcementRLIR} & $35.19$ & $11.23$ & $2.72$ & $1.93$ & $1.52$ & $0.12$ & $0.33$ \\
    Q-chunking \cite{li2025reinforcementchunk} & $39.70$ & $13.90$ & $1.94$ & $1.65$ & $0.80$ & $0.09$ & $0.22$ \\
    RAD \cite{gao2025rad} & $49.24$ & $12.85$ & $1.68$ & $2.69$ & $0.68$ & $0.11$ & $0.19$ \\
    \textbf{Ours} & $\textbf{52.97}$ & $\textbf{13.98}$ & $\textbf{1.56}$ & $\textbf{3.59}$ & $\textbf{0.52}$ & $\textbf{0.08}$ & $\textbf{0.11}$ \\
    \bottomrule
\end{tabular}}
\label{table:close}
\vspace{-3mm}
\end{table*}

\begin{figure}
    \centering
    \includegraphics[width=0.96\linewidth]{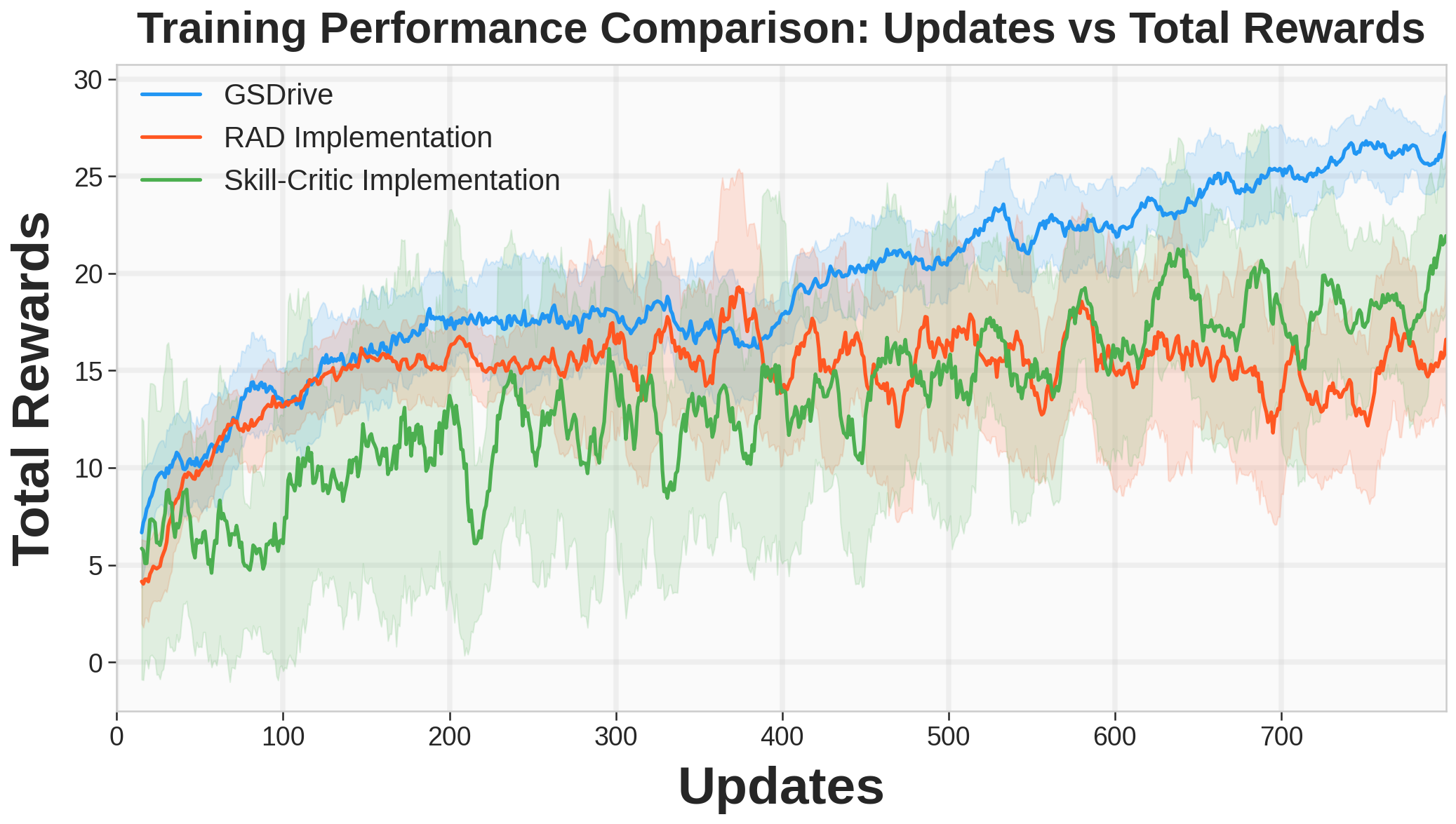}
    \caption{The training performance comparison, where the shaded area represents the rolling standard deviation of the mean rewards.}
    \label{fig:compare}
\vspace{-5mm}
\end{figure}

\subsection{Training Performance}

We show the comparison of the training process of $600$ scenes between our GSDrive and other RL-based methods in Fig. \ref{fig:compare}. Each update collects the mean rewards from 4 parallel environments running 8 steps each, with trajectory probing in $K=5$ modes and $H=6$ horizons. In RAD \cite{gao2025rad}, the 3DGS environment is also used for policy learning. However, the critical reward signals, such as collisions, are based on current environmental interactions. When a collision happens, it results from a sequence of poor decisions made over several seconds, not just the final unsafe action. The policy only receives a significant reward signal at the exact moment a critical event occurs, which may lead to premature convergence during training on a suboptimal policy with lower overall performance. Shifting from this base, we follow Skill-Critic \cite{hao2024skill}, encoding actions from our dataset into continuous latent variables. Then, hierarchical policy learning with both discrete and continuous action outputs is conducted in the 3DGS environment. The joint optimization on a hybrid action space introduces inherent instability in the Actor-Critic framework, leading to noisy training signals.

\begin{figure*}[h!]
    \centering
    \includegraphics[width=0.96\linewidth]{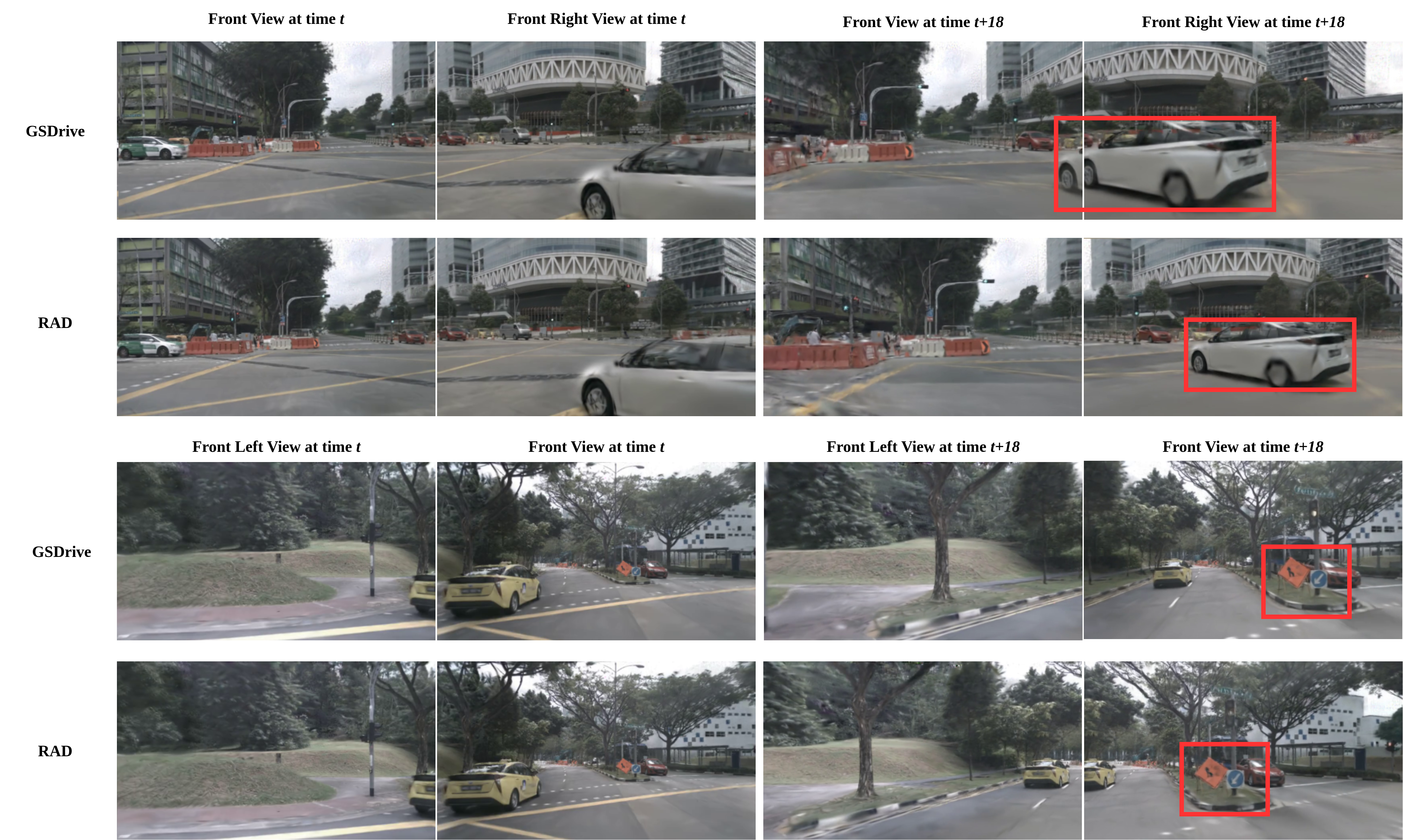}
    \caption{Qualitative comparisons in the closed-loop test.}
    \label{fig:vis}
\vspace{-5mm}
\end{figure*}

Our GSDrive, instead of waiting for a sparse and catastrophic reward, can simulate the likely future trajectory and evaluate it for safety, comfort, and progress. It can then provide a dense, informative reward signal back to the RL training immediately. Optimizing only over a discrete set of actions, our RL objectives lead to more stable and superior training performance than other RL methods.

\subsection{Closed-loop Evaluation}

To perform a comprehensive evaluation of all the methods, we define the following closed-loop metrics for our 3DGS environment:
\begin{itemize}
    \item Episode Reward (ER $\uparrow$) $r_{\text{episode}}$
    \item Driving Speed [$m/s$] (DS $\uparrow$) $v_{\text{episode}}$
    \item Maximum Acceleration [$m/s^2$] (MA $\downarrow$) $a_{\text{episode}}$
    \item Lane Changes (LC $\uparrow$) $n_{\text{episode}}$
    \item Maximum Action Jerks [$m/s^3$] (MAJ $\downarrow$) $j_{\text{episode}}$
    \item Maximum Steering Angle [$rad$] (MSA $\downarrow$) $\delta_{\text{episode}}$
    \item Collision Rate (CR $\downarrow$) $\lambda_{\text{episode}}$
\end{itemize}
where the acceleration $a_{t}^{\text{episode}} = \frac{v^{\text{episode}}_{t+1}-v^{\text{episode}}_t}{\Delta t}$, the jerk $j^{\text{episode}}_t = \frac{a^{\text{episode}}_{t+1} - a^{\text{episode}}_t}{\Delta t}$, and the steering angle $\delta^{\text{episode}}_t = \arctan2(\Delta y_t, \Delta x_t)$.

In Tab. \ref{table:close}, we show the average results across the $50$ test episodes for $200$ scenes. For RLIR \cite{ye2025reinforcementRLIR}, we leverage an Inverse Dynamics Model (IDM) to recover actions from real-world datasets rather than relying on original world model generations to construct reward signals. Yet, the performance of the resulting RL policy is intrinsically capped by the accuracy of the IDM, coupled with extensive IDM training. Following Q-chunking \cite{li2025reinforcementchunk}, we extend the original action space to a temporally chunked action space, relying on a fixed action chunk length. However, the actions are still executed in an open-loop manner in the environment, enforcing exploration via temporally coherent chunks from offline data. 

Consequently, our GSDrive demonstrates the best test performance through future-aware reward shaping from the 3DGS environment. Notably, our approach achieved the highest episode reward while simultaneously maintaining high-speed traversal capabilities, highlighting the effectiveness of leveraging trajectory probing feedback for driving tasks.

\subsection{Qualitative Results}

In Fig. \ref{fig:vis}, we show the multi-view qualitative comparisons between our GSDrive and RAD in the closed-loop test sequences, where we mark the same objects with red rectangles to illustrate the differences in the executed actions. In the scenarios of straight yielding while avoiding out-of-distribution obstacles, our GSDrive maintains a safe distance from both agent vehicles and obstacles amid the pursuit of its forward driving goal.

\subsection{Ablation Study}

We conduct the ablation study with training and testing on the random subset of $200$ scenes in our dataset. In Tab. \ref{table:traj}, to maintain the trajectory head with the ability for multi-mode prediction during trajectory probing, we compare the DDPM and DDIM under the diffusion model with our methods \cite{ho2020ddpm, song2020ddim}. For the experiments, we use $20$ denoising steps to investigate the low-step regime for the driving task, where diffusion models encounter difficulties in learning curved vector fields. While flow matching is able to predict smooth and physically consistent trajectories at low steps \cite{lipman2022flow}, it suffers from mode averaging problems, with generated trajectories tending to be fuzzy or conservative, lacking the decisive edges of driving. Thus, CFG benefits this mission by providing strict adherence to the fused conditional queries, ensuring stable trajectory generation that is both physically consistent and precisely aligned with the multiple modes \cite{ho2022classifier}.

In Tab. \ref{table:self}, we use a direct MLP-based action head to predict the action logits from the fused queries, compared to our combination of trajectory-constructed and residual logits. With the probing signals from the trajectory priors, the combined logits establish a hierarchical optimization objective, outperforming a direct MLP that lacks this structural guidance and must solve the entire action space mapping in a single transformation.

\begin{table}[h]
\vspace{-2mm}
\centering
\caption{Ablation study on trajectory head.}
\label{table:traj}
\resizebox{0.48\textwidth}{!}{
\begin{tabular}{c|c c c c c c}
\toprule
     Methods & ER $\uparrow$ & DS $\uparrow$ & MA $\downarrow$ & MAJ $\downarrow$ & MSA $\downarrow$ & CR $\downarrow$ \\ \hline
    DDPM \cite{ho2020ddpm} & $45.08$ & $9.34$ & $2.07$ & $1.21$ & $0.18$ & $0.24$ \\
    DDIM \cite{song2020ddim} & $46.32$ & $9.69$ & $2.03$ & $1.04$ & $0.13$ & $0.22$ \\
    Flow Matching \cite{lipman2022flow} & $51.77$ & $9.80$ & $2.03$ & $0.99$ & $0.11$ & $0.22$ \\
    Flow Matching + CFG \cite{ho2022classifier} & $\textbf{51.82}$ & $\textbf{9.85}$ & $\textbf{2.01}$ & $\textbf{0.92}$ & $\textbf{0.10}$ & $\textbf{0.19}$ \\
    \bottomrule
\end{tabular}}
\vspace{-3mm}
\end{table}

In probing rewards, sampled $K$ modes and $H$ future steps balance exploration quality against computational efficiency. This experiment presents a rigorous ablation study to systematically investigate their influence, where we evaluate five levels of $K$, spanning from deterministic selection ($K=1$) to high exploration ($K=10$) and $H$, determining the temporal extent of credit assignment.

\begin{table}[h]
\vspace{-1mm}
\centering
\caption{Ablation study on direct action head, residual action head, and trajectory probing rewards.}
\label{table:self}
\resizebox{0.48\textwidth}{!}{
\begin{tabular}{c c c|c c c c c c}
\toprule
     \makecell{Direct \\ Action Head} & \makecell{Residual \\ Action Head} & \makecell{Probing \\ Rewards} & ER $\uparrow$ & DS $\uparrow$ & MA $\downarrow$ & MAJ $\downarrow$ & MSA $\downarrow$ & CR $\downarrow$ \\ \hline
     & \CheckmarkBold & & $40.21$ & $8.22$ & $2.87$ & $0.74$ & $0.13$ & $0.23$ \\
     
    \CheckmarkBold & & \CheckmarkBold & $43.09$ & $9.69$ & $2.23$ & $0.70$ & $0.14$ & $0.20$ \\

     & \CheckmarkBold & \CheckmarkBold & $\textbf{49.08}$ & $\textbf{10.20}$ & $\textbf{2.09}$ & $\textbf{0.66}$ & $\textbf{0.13}$ & $\textbf{0.16}$ \\
    \bottomrule
\end{tabular}}
\vspace{-2mm}
\end{table}

In Tab. \ref{table:K}, we show the Collision Rate alongside the Success Rate, which is the percentage of scenes in which the vehicle successfully completes the planned route without any failures. In the test set of $50$ scenes, with fixed $H=6$, increasing $K$ improves driving performance up to a saturation point, beyond which marginal gains diminish due to mode redundancy.

\begin{table}[h]
\vspace{-2mm}
\centering
\caption{Collision Rate ($\downarrow$) and Success Rate ($\uparrow$) under varying $K$ modes.}
\label{table:K}
\begin{tabular}{|c|c|c|c|c|c|}
\hline
$K$ & 1 & 3 & 5 & 7 & 10 \\ \hline
Collision Rate & $12.3\%$ & $7.8\%$ & $4.9\%$ & $5.2\%$ & $5.0\%$ \\ \hline
Success Rate & $78.2\%$ & $85.6\%$ & $92.3\%$ & $92.1\%$ & $91.8\%$ \\ \hline
\end{tabular}
\vspace{-2mm}
\end{table}

Extended horizons provide richer reward signals for temporal credit assignment, enabling the policy to reason about long-term consequences. However, as shown in Tab. \ref{table:H}, under $K=5$, the shadow-environment probing accumulates prediction errors as the horizon extends, leading to diminished accuracy of the reward signal.

\begin{table}[h]
\vspace{-2mm}
\centering
\caption{Collision Rate ($\downarrow$) and Success Rate ($\uparrow$) under varying $H$ horizons.}
\label{table:H}
\begin{tabular}{|c|c|c|c|c|c|}
\hline
$H$ & 1 & 3 & 6 & 10 & 15 \\ \hline
Collision Rate & $11.2\%$ & $6.4\%$ & $4.9\%$ & $6.1\%$ & $7.8\%$ \\ \hline
Success Rate & $80.1\%$ & $88.7\%$ & $92.3\%$ & $89.4\%$ & $86.2\%$ \\ \hline
\end{tabular}
\vspace{-2mm}
\end{table}

%% file: Con.tex
We presented GSDrive, which reinforces E2E driving policies through multi-mode trajectory probing within a 3D Gaussian Splatting environment. Our approach pretrains the policy via IL with an Optimal Transport-guided flow matching head. During RL, this predictor acts as a prospective reward probe, where the candidate trajectories are rolled out in the 3DGS simulator to evaluate their future physical consequences. We found that the proposed trajectory probing enables the policy to foresee the outcomes of current decisions, allowing preemptive correction to avoid potential hazards or task failures. By dynamically balancing trajectory-constructed logits with residual logits across training stages, our framework reconciles imitation priors with exploratory RL, enabling deviation from demonstrator behavior when safety demands while preserving critical inductive biases.

Our framework naturally supports lifelong learning, where unlabeled observations can be integrated into 3DGS reconstruction without manual annotation. Future-aware probing feedback from 3DGS proves highly effective for continuous policy improvement. The bidirectional IL-RL knowledge transfer, mediated by the trajectory probe, substantially outperforms methods relying solely on instantaneous environmental signals or inverse dynamics models.